\documentclass[a4paper,twocolumn]{article} 

\ifx\pdfoutput\undefined
   \usepackage[dvips]{graphicx}
   \DeclareGraphicsExtensions{.eps} 
 \else
   \usepackage{graphicx}
   \DeclareGraphicsExtensions{.pdf,.jpg,.png,.mps} 
\fi
\graphicspath{{./../Figures/},{./../Proposal_PhD_Chang_FFPD/Figures/}, {./../sn_version/}}

\usepackage{amsmath,amssymb}   
\usepackage[ansinew]{inputenc} 
\usepackage[english]{babel}    
\usepackage[round,authoryear]{natbib}  
\usepackage[font=footnotesize,labelfont=bf]{caption}
\usepackage{algorithm}
\usepackage{algpseudocode}
\usepackage{titlesec}

\titleformat*{\section}{\Large\bfseries}
\titleformat*{\subsection}{\large\bfseries}
\titleformat*{\subsubsection}{\normalsize\bfseries}
\titleformat*{\paragraph}{\normalsize\bfseries}

\usepackage{hyperref}
\usepackage{tabularx}
\newcolumntype{L}[1]{>{\raggedright\let\newline\\\arraybackslash\hspace{0pt}}p{#1}}
\newcolumntype{C}[1]{>{\centering\let\newline\\\arraybackslash\hspace{0pt}}p{#1}}
\newcolumntype{R}[1]{>{\raggedleft\let\newline\\\arraybackslash\hspace{0pt}}b{#1}}
\usepackage{multirow}
\newcolumntype{P}[1]{>{\raggedright\arraybackslash}m{#1}}
\usepackage[a4paper, total={6.5in, 10in}]{geometry}

\usepackage{cuted}

\usepackage{xstring} 

\usepackage[colorinlistoftodos,prependcaption,textsize=tiny]{todonotes}
\usepackage{cleveref}

\title{{\LARGE \textbf{Unsupervised Skin Feature Tracking with Deep Neural Networks}}\\
\vspace*{0.3cm}
{\large Jose Ramon Chang, MSc \textsuperscript{1} and Torbj{\"o}rn E. M. Nordling, MSc, PhD \textsuperscript{1,2}\footnote{Corresponding author}}\\
\vspace*{0.2cm}
{\footnotesize \begin{tabular}{C{6cm} C{0.1cm} C{7cm}}
\textsuperscript{1}Department of Mechanical Engineering& &\textsuperscript{2}Department of Applied Physics and Electronics\\
National Cheng Kung University& &Ume\aa{} University\\
Tainan 701, Taiwan & &90187 Ume\aa, Sweden \\
\end{tabular}}\\
\vspace*{10pt}
{\footnotesize \texttt{\{jose.chang, torbj{\"o}rn.nordling\}@nordlinglab.org}}
}
\date{\vspace*{-1.6cm} } 

\renewenvironment{abstract}
{
  \centerline{\large \bfseries \scshape Abstract}
  \vspace*{0.2cm} \footnotesize 
}
{

}
\begin{document}               
\maketitle
\abstract{\itshape
Facial feature tracking is essential in imaging ballistocardiography for accurate heart rate estimation and enables motor degradation quantification in Parkinson's disease through skin feature tracking. 
While deep convolutional neural networks have shown remarkable accuracy in tracking tasks, they typically require extensive labeled data for supervised training.
Our proposed pipeline employs a convolutional stacked autoencoder to match image crops with a reference crop containing the target feature, learning deep feature encodings specific to the object category in an unsupervised manner, thus reducing data requirements.
To overcome edge effects making the performance dependent on crop size, we introduced a Gaussian weight on the residual errors of the pixels when calculating the loss function. 
Training the autoencoder on facial images and validating its performance on manually labeled face and hand videos, our Deep Feature Encodings (DFE) method demonstrated superior tracking accuracy with a mean error ranging from 0.6 to 3.3 pixels, outperforming traditional methods like SIFT, SURF, Lucas Kanade, and the latest transformers like PIPs++ and CoTracker.
Overall, our unsupervised learning approach excels in tracking various skin features under significant motion conditions, providing superior feature descriptors for tracking, matching, and image registration compared to both traditional and state-of-the-art supervised learning methods.
}


\section{Introduction}\label{sec:introduction}
\label{sec:intro}
Distinctive facial feature points, typically found around the eyes, nose, chin, and mouth, encapsulate the most pertinent information for both discriminative and generative purposes \citep{Wang2018}. 
With the proliferation of large and readily available facial imagery datasets, researchers have increasingly turned to deep learning techniques to enable computers to undertake various tasks. 

Most state-of-the-art deep learning methods have been trained using a supervised learning paradigm \citep{Goodfellow2016}.
Here, the parameters of the model are updated based on the error of its prediction and the ground truth, $i.e.$ label.
Curating a labelled dataset is typically one of the hardest parts in the pipeline of a deep learning application because of the sample quality, sample quantity, labelling complexity, class imbalance, data privacy issues, data bias, and label certainty. 
For optical flow, it is especially difficult to obtain datasets of real objects because the labelling is very intensive.
For a single image, a large number of points must be labelled with a precision that is hard for a human to achieve when labelling not well-defined salient, features. 
This is why the datasets for training of dense optical flow are synthetic. 
We found 11 synthetic datasets, such as FlyingThings3D \citep{mayer2016large}, Max Planck Institute (MPI) Sintel \citep{butler2012naturalistic}, Virtual Karlsruhe Institute of Technology and Toyota Technological Institute (KITTI) \citep{gaidon2016virtual}, PointOdyssey \citep{zheng2023pointodyssey}, Tracking Any Point (TAP)-Vid \cite{doersch2022tap}, where the motion and positions of points are pre-defined and no manually labelled ones. 
We manually labelled skin features in facial and hand videos for remote photoplethysmography and the postural tremor test of the Movement Disorder Society revised Unified Parkinson's Disease Rating Scale.

Tracking the position of a specific skin feature poses significant challenges. 
Skin in images often appears uniform, with only sporadic locally distinctive features. 
The patterns of these distinct features may undergo considerable deformation due to changes in facial expression or lighting conditions. 
While some skin features, such as moles or birthmarks, are easily recognizable by the human eye, others are not as discernible. 
We selected moles and nose tips as our features to be tracked. 
While deep learning has gained significant traction in medical applications related to skin, including remote photoplethysmography (rPPG) \citep{ni2021review, cheng2021deep}, gaze estimation \citep{cheng2021appearance}, speech recognition \citep{lee2021biosignal}, skin lesion segmentation \citep{stofa2021skin}, and driver fatigue detection \citep{sikander2018driver}, there has been limited original research explicitly quantifying the matching error of skin features. 
Manni et al. matched keypoints on patients' backs using hyperspectral images \citep{manni2020hyperspectral}. 
Employing a hyperspectral camera, they captured spectral bands and applied the Saliency-Band-Based Selection algorithm (SBBS) to reduce dimensions. 
Utilizing Speeded Up Robust Features (SURF) and DEep Local Feature (DELF) descriptors, they achieved a localization error of 0.25 mm with DELF outperforming SURF.
In this study, we compared the performance of SURF, Scale Invariant Feature Transform (SIFT), Lucas-Kanade (LK), Persistent Independent Particles (PIPs)++, CoTracker, and a stacked autoencoder, called Deep Feature Encodings (DFE), trained using two different loss functions.

Most deep learning feature tracking research builds upon object detection.
Object detection networks output bounding boxes defined by $x$ and $y$ coordinates, box dimensions, confidence scores, and class probabilities. 
Multiple Object Tracking (MOT) assigns bounding boxes to detected objects, associating each with a unique target ID \citep{Ciaparrone2020}. 
This ``tracking-by-detection" approach iteratively detects objects across video frames and associates bounding boxes between successive frames. 
Faster-Region based Convolutional Neural Networks (RCNN) \citep{Ren2015}, Single Shot Detector \citep{Liu2016ssd}, You Only Look Once (YOLO) \citep{reis2023real}, Mask RCNN \citep{He2017}, and Region-based Fully Connected Networks (R-FCN) \citep{Dai2016} excel at accurate object detection, challenges arise when tracking small objects with low resolutions, such as skin features. 
Variational approaches like Deepflow \citep{weinzaepfel2013deepflow} and FlowNet \citep{dosovitskiy2015flownet} estimate dense optical flow using CNNs, DELF \citep{noh2017large} calculates feature descriptors and keypoint selections, and Detector-Free Local Feature Matching with Transformers (LoFTR) \citep{sun2021loftr} leverages transformers for dense matching between keypoints. 
While these methods offer efficient matching capabilities, our focus lies in learning feature representations rather than precise keypoint association.
Since autoencoders offer an alternative approach, sidestepping the need for extensive data labelling required by supervised learning methods, we trained one.

Introduced in 2023, PIPs++ advances long-term fine-grained tracking through the introduction of PointOdyssey, a synthetic dataset and framework designed for training algorithms. 
The realistic animation of PointOdyssey and diverse scene construction enable superior training outcomes, surpassing existing methods. 
The enhancements to the PIPs tracking technique and dataset significantly contribute to the progress of long-term tracking algorithms. 
Notably, PIPs++ demonstrates superior average position accuracy compared to RAFT \citep{teed2020raft}, DINO \citep{caron2021emerging}, and TAP-net \citep{doersch2022tap} on the TAP-Vid DAVIS \citep{doersch2022tap} and CroHD \citep{sundararaman2021tracking} datasets. 
Additionally, CoTracker, a transformer-based model developed by MetaAI in late 2023, excels in accuracy and robustness by jointly tracking dense points across video frames. 
Its innovative features, including virtual tracks, enable simultaneous tracking of 70 thousand points, ensuring superior performance for short and long-term tasks. 
The effectiveness of CoTracker is evident in both qualitative and quantitative benchmarks, exhibiting impressive tracking results even in challenging scenarios such as occlusions. 
Notably, CoTracker outperforms other state-of-the-art deep learning methods in terms of 3-pixel accuracy on the BADJA \citep{biggs2019creatures} and FastCapture \citep{rocco2023real} datasets.

For a more in-depth review of object and feature tracking methods, we refer to our previous work \citep{chang2021skin}.
In the upcoming sections, we will characterize our dataset, describe other state-of-the-art deep learning models we compared to, and introduce a weighted variant of our DFE. 
Finally, we will assess the performance of feature matching and tracking, while also presenting the fitness landscape of SIFT, SURF, DFE, and wDFE.

\section{Materials and Methods}
\label{sec:methods}

\begin{figure*}[tb!]
\centering
\includegraphics[width=\textwidth]{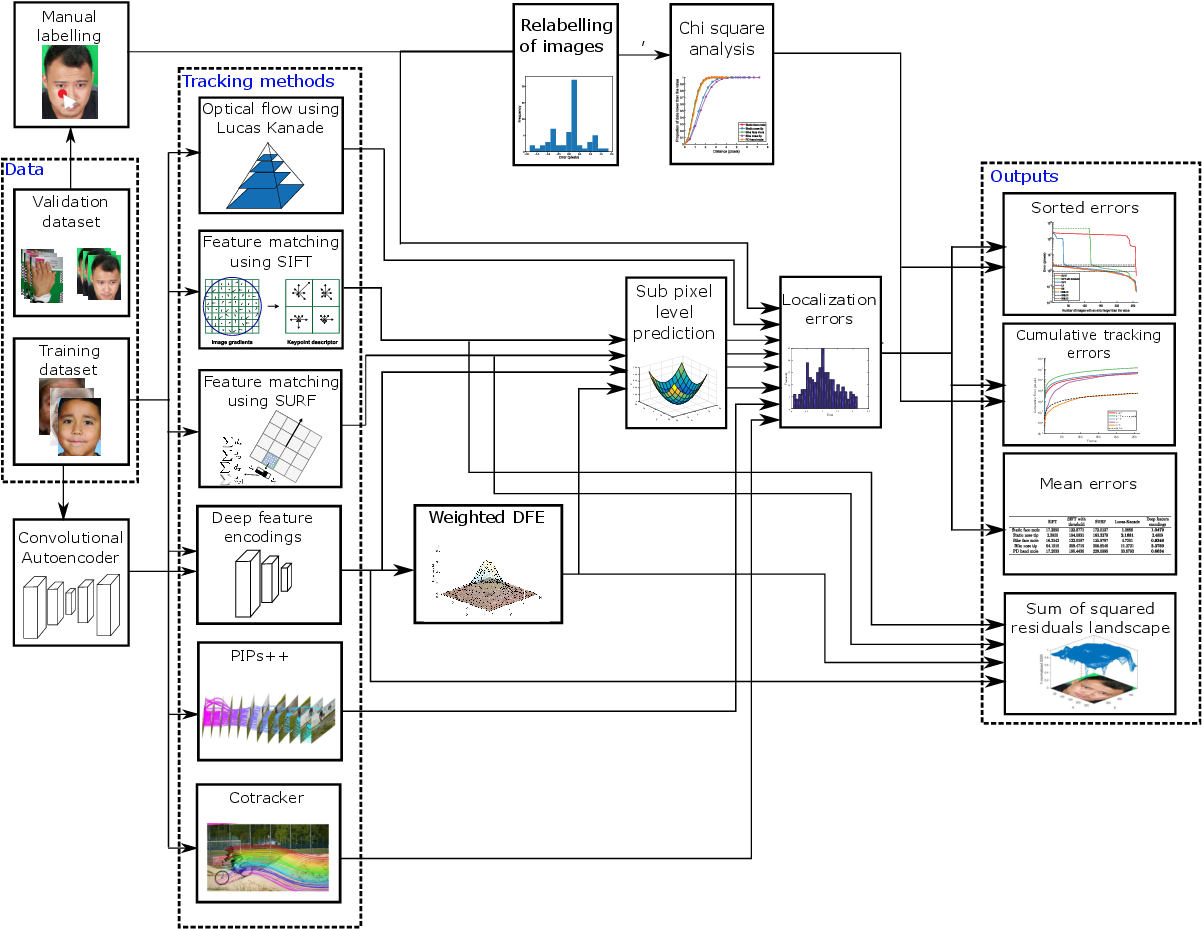}
\caption[Schematic workflow of our analyses and experiments.]{Schematic workflow of our analyses and experiments. The UTKface (training) dataset was used to train the autoencoder used in our DFE method. Our validation dataset was manually labelled to obtain the ground truth of the location of the features. The feature tracking methods were used to predict the localisation of the skin features in the validation dataset. For SIFT, SURF, DFE, and wDFE the optimal match is determined to obtain subpixel level predictions before calculating the error relative to the manual labelling. Based on those errors; the sorted errors, cumulative tracking errors, and mean errors are reported. The SSR landscape takes the high-dimensional representations of the points as input and visualise them.}
\label{fig:ffpd_flowchart} 
\end{figure*}

\begin{figure}[tb!]
\centering
\includegraphics[width=\linewidth]{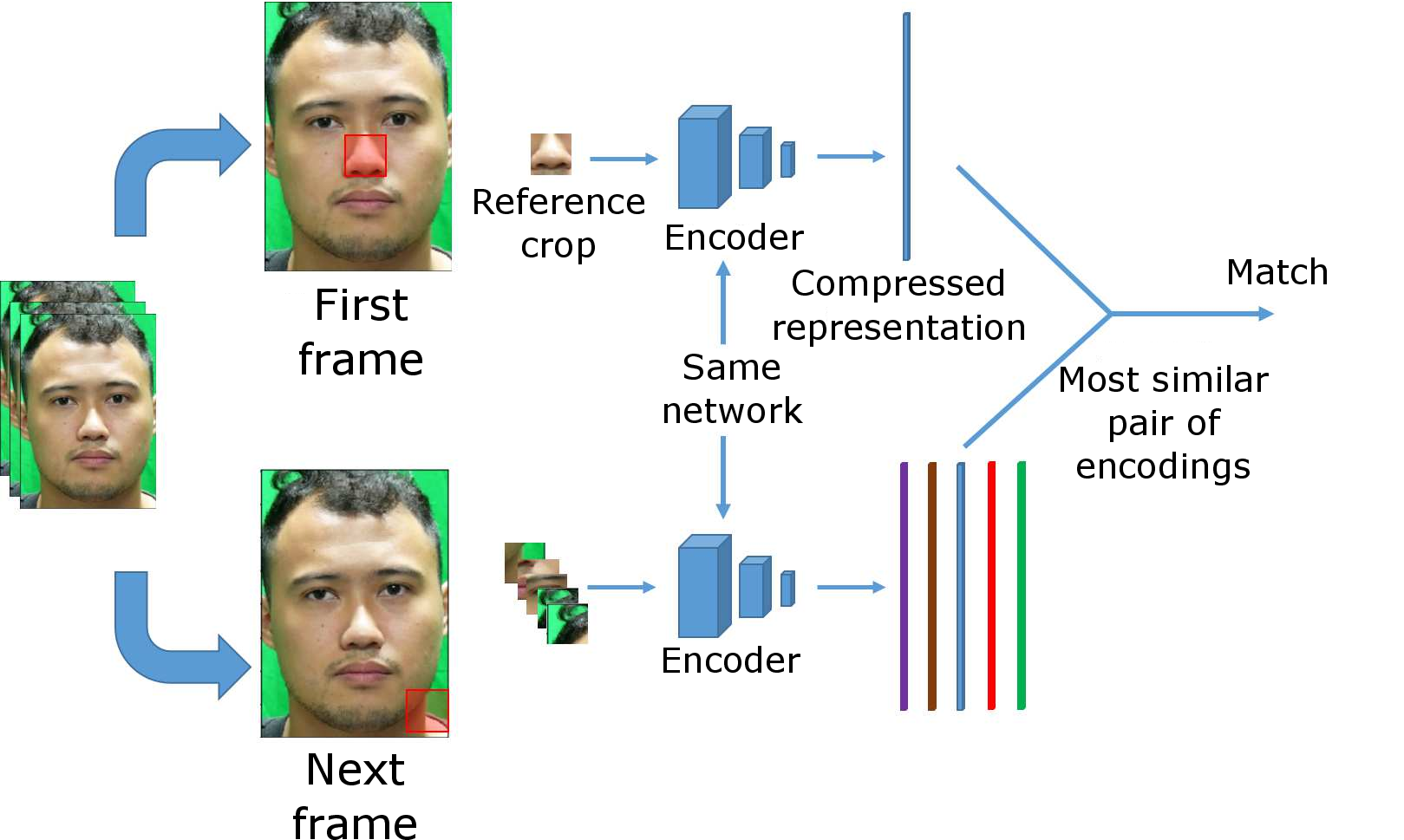}
\caption[Flowchart of the algorithm for using an autoencoder for matching face features.]{Flowchart of the algorithm for using an autoencoder for matching facial features.}
\label{fig:Autoencoder_matching_algorithm_flowchart}
\end{figure}

Tracking features in a video involves successive feature matching across all frames. 
We compare the feature description capability and matching performance of various algorithms: SIFT, SURF, LK, PIPs++, CoTracker, and our DFE. 
For an overview of our experimental setup, see Figure~\ref{fig:ffpd_flowchart}. 
We have selected two facial features under different motion conditions and one hand feature of a patient diagnosed with Parkinson's Disease (PD) during a postural tremor test. 
The chosen facial features are a mole and the tip of the nose, while the hand feature is a mole, all consistently visible in every frame of our validation dataset. 
The objective is to locate a reference point from the initial frame in each subsequent frame. 
In matching for SIFT, SURF, and DFE, we minimize the SSR between predicted and reference descriptors. 
The matching process of DFE is detailed in Figure~\ref{fig:Autoencoder_matching_algorithm_flowchart}. 
LK determines matches by minimizing optical flow equations. 
PIPs++ and CoTracker rely on their models, using color information encoded in RGB space, while DFE employs CIELAB color space. 
SIFT, SURF, and LK operate on grayscale images.
To evaluate the matching error of the methods, the images were manually labeled with pixel-level accuracy once.

A Chi-square analysis was conducted to set a threshold distinguishing errors from human labeling and those from the tracking method. 
For each body part and motion condition, 15 images per skin feature were relabeled five times, spaced over three months to prevent consecutive exposure, resulting in 90 samples per condition. 
Errors in $x$ and $y$ coordinates were compared against the mean of all labeling attempts, yielding normal error distributions. 
The standard deviations of the error distributions for the manual relabeling attempts are in Table 1 in our previous work \citep{chang2021skin}.
The Chi-square test statistic, $\hat{\chi}^2$, was computed and evaluated against a Chi-square distribution with $2n$ degrees of freedom. 
Nose tips exhibited larger error standard deviations than moles due to their lack of distinctiveness.

\subsection{Training dataset}
The autoencoder of our DFE method is trained on the University of Tennessee, Knoxville Face (UTKFace) dataset \citep{Zhang2017age}, featuring diverse RGB face images spanning ages from 0 to 116 years, converted to CIELAB color space CIELAB color space \citep{mclaren1976xiii}, for perceptual uniformity. 
With variations in backgrounds, lighting, and facial attributes, including makeup and accessories, the dataset presents a rich array of facial features. 
To train a skin feature-specialized autoencoder, we extract aligned face crops from the dataset using an OpenCV ResNet-10 model, and cropped patches of $31 \times 31$ resulting in approximately 1.9 million training crops.
A more detailed description of the data preprocessing can be found in our previous work \citep{chang2021skin}.

\subsection{Validation dataset}

The videos utilized for testing our algorithms were gathered within our laboratory for two distinct projects. 
These datasets can be categorized as follows:

\textbf{Remote ballistocardiography:} This dataset, recorded in full HD resolution ($1920 \times 1080$ pixels) at 50 FPS was obtained to estimate heart rate remotely by tracking subtle head movements caused by blood flow through the carotid artery. The specifics of the examination procedure are outlined within our protocol \citep{Vivaldy2023}.

These videos were recorded under two motion conditions:

\textbf{Small motion--}The subject is instructed to simply look at the camera for a couple of minutes without moving, referred to as static condition.

\textbf{Large motion--}The subject is filmed while riding an exercise bicycle. 
The motion is larger than in the static case and periodic. 
This is referred to as the bike condition.

\textbf{Parkinson's disease postural tremor test:}
This dataset comprises videos of hands from Parkinson's disease (PD) patients, recorded to digitize the Unified Parkinson's Disease Rating Scale (UPDRS) motor exams. We use these videos to evaluate our method's effectiveness in tracking skin features beyond the face. Details of the examination procedure can be found in our protocol \citep{ashyani2022digitization}. 
The videos depict the left hand of the subjects, selected from available views (front, right, left), and were recorded at 240 FPS in HD resolution (1280 x 720 pixels).
This video is denoted as Postural Tremor (PT) test.

Both video datasets were initially encoded in the RGB color space but were later transformed into the CIELAB color space. 
Additionally, the videos underwent subsampling to 2 FPS, which increased the displacement between frames, consequently elevating the tracking complexity while reducing computation time. 
Following temporal subsampling, the face videos comprised 261 frames for the static condition and 170 frames for the bike condition, while the hand video contained 40 frames. 
To align with images from the UTKFace dataset and streamline computational evaluation, all images were resized to $420\times 300$ pixels using inter-area interpolation in the RGB color space. 
Manual annotation of the face mole, nose tip, and hand mole was performed in all frames at the $420\times 300$ resolution.

\subsection{Implementation of traditional computer vision techniques}

SIFT \citep{Lowe2004} and SURF \citep{Bay2006}, although not cutting-edge, serve as foundational benchmarks for our research, inspiring our approach to feature matching. 
Our method, like SIFT and SURF, requires no training data, making it suitable for datasets with limited labels. 
Additionally, LK \citep{lucas1981iterative} offers efficiency, simplicity, and robustness in sparse optical flow tasks. 
Together, these methods are widely recognized in the computer vision community, we will therefore focus on expanding on other state-of-the-art approaches.
For a more in-depth description of these traditional methods, we refer to our previous work \citep{chang2021skin}.

\textbf{SIFT:} We used the default implementation of SIFT in OpenCV 3.4.2.17, $i.e.$ the $128$-feature descriptor.
The parameters used were \texttt{nfeatures = 0}, \texttt{nOctaveLayers = 3}, \texttt{contrastThreshold = 0.04}, \texttt{edgeThreshold = 10}, and $\sigma = 1.6$.

\textbf{SURF:} We have used the extended version of SURF as implemented in OpenCV 3.4.2.17 to have 128 features and match the dimensionality of SIFT descriptors.
The parameter values are \texttt{hessianThreshold = 100}, \texttt{nOctaves = 4}, \texttt{nOctaveLayers = 3}, \texttt{extended = true}, and \texttt{upright = false}.

\textbf{Lucas-Kanade:} We used the implementation from OpenCV 3.4.2.17 with a pyramid of 5 levels, $L_m=4$, with a $10 \times 10$ observation window at each level. Skin is rather homogeneous with distinct features in some places.
We, therefore, want to be able to select an area of distinct features to track, which is why it makes sense to work with patches.

\subsection{PIPs++}

Persistent Independent Particles++ (PIPs++) \citep{zheng2023pointodyssey} improves long-term tracking and establishes a fine-grained tracking standard, expanding upon PIPs \citep{harley2022particle}. 
PIPs++ utilizes a dataset containing 104 lengthy videos with an average duration of 2,000 frames. 
It takes an 8-frame RGB video as input $V$, along with a coordinate $P_1=(x_1, y_1)$ indicating a target to track and outputs the tracks $\hat{P}_t=(\hat{x}_t, \hat{y}_t)$ and visibility annotations $v_t$, $i.e.$ fuzzy values that refer to whether a point is visible or occluded in a given frame.
To ensure continuous tracking beyond the initial 8-frame period, a chaining strategy is employed, which involves re-initializing the tracker at the last frame with adequate visibility. 

The model operates in two stages: initialization and iterative updates.
During initialization, a 2D residual ConvNet \citep{He2016deep} is used to compute feature maps of the reference feature.
The iterative update stage aims to refine position estimates through similarities between feature maps, followed by updates to the sequence of positions and features using a 1D ResNet to widen the temporal view to enhance accuracy and speed on long videos.
It also utilizes a template-update mechanism to adapt to appearance changes.

For implementation, default parameters from the model provided on their project page (\url{https://pointodyssey.com/}) were used.
Given that the face mole and nose tip are not adjacent, they were tracked individually, as their implementation only a uniform grid of adjacent points to be simultaneously tracked through parallel computation.

\subsection{CoTracker}
Cotracker \citep{karaev2023cotracker} is an algorithm that takes an input video $V$ and starting locations $(P_{t^i}^i, t^i)_{i=1}^N$ of multiple $N$ keypoints and outputs the predictions $\hat{P}_t^i = (\hat{x}_t^i, \hat{y}_t^i)$ and visibility flags $v_t^i$ for all valid times.

This method uses a transformer network trained on TAP-Vid Kubric \citep{doersch2022tap} and PointOdyssey \citep{zheng2023pointodyssey} for improving track predictions in a video sequence. 
CoTracker can track 70 thousand points jointly and simultaneously. 
Tracks are represented as a grid of input tokens that capture position, visibility, appearance, and correlation features. 
Image features are extracted using a convolutional neural network, and track features are initially sampled from these features and updated by the network.
The correlation features facilitate matching tracks to images. 
The transformer is applied multiple times, iteratively improving the track predictions. 
The final visibility mask is sigmoidally updated and initialization involves sinusoidal positional encoding based on the starting position of the keypoints.

Cotracker can process long videos by creating smaller overlapping segments, $ i.e.$ windows.
The video $V$ with length $T'$ is split into $J=\frac{2T'}{T-1}$  windows of length $T$ with a half-window overlap.
The transformer is applied $M$ times to each window, and the output of one window serves as the input for the next allowing tracking points throughout the entire video. 
To properly deal with these semi-overlapping windows, the transformer iteratively updates for each window in an unrolled manner, $i.e.$ unrolled learning.

For CoTracker, the default parameters provided on their project page (\url{https://co-tracker.github.io/}) were used. 
Additionally, simultaneous tracking of the face mole and nose tip was performed under various motion conditions, while individual tracking of the hand mole was conducted for the Parkinson's Disease (PD) dataset, which only included data for this feature.

\subsection{Deep feature encodings}

DFE employs unsupervised deep learning to enhance skin feature representations for improved tracking, addressing the scarcity of labeled samples in our datasets. 
The model consists of an autoencoder comprising an encoder subnetwork compressing the input data into a latent space and a decoder subnetwork reconstructing the input from the encoded representation. 
This network is formulated as $y(x)=g(f(x))$, where $f$ and $g$ are constructed using 2D convolutional and transpose convolutional layers. 
For a full description of the architecture of the autoencoder, we refer to Figure 4 in our previous work \citep{chang2021skin}.
To ensure a fair comparison with SIFT and SURF, the latent space of the autoencoder is set to 128 features, yielding a compression factor of 0.0444, with the Mean Squared Error (MSE) loss function employed for training. 
The architecture involves symmetrical convolutional layers, Batch Normalization, ReLU nonlinearity, and a decoder network mirroring the encoder. 
Training utilized the Adamax optimizer, with the network comprising 1,160,847 trainable parameters, trained for 1000 epochs to a training MSE of $9.4 \times 10^{-5}$ in approximately 138.9 hours on a system with three ASUS 1080 Ti GPUs.

\subsubsection{Latent space encoding}

For each image $I_{x,y}$ we select crops $C(i,j)$ of window size $w = (w_x, w_y) = (31,31)$ with stride 1 for all possible positions $(i,j)$ in $(x,y)$ such that 
\begin{align*}
x_{min}&=floor(i-w_x/2)\\
x_{max}&=floor(i+w_x/2)\\
y_{min}&=floor(j-w_y/2)\\
y_{max}&=floor(j+w_y/2)\\
C(i,j) &= I[x_{min}:x_{max},y_{min}:y_{max},:].
\end{align*}
Note that crops can only be obtained for positions where $floor(w_x/2)+1<i<I_w-floor(w_x/2)$, and $floor(w_y/2)+1<j<I_h-floor(w_y/2)$ with $I_w$, $I_h$ representing the width and height of image $I$.
For positions where a crop can not be obtained, we set the value of that position as 0 in all our maps.

These crops $C(x,y)$ are then converted into encodings $h(x,y)$ by projecting these into the 128-dimensional latent space using the encoder subnetwork of DFE $i.e.$ encoding function $f$, such as
\begin{align*}
h(x,y)=f(C(x,y)).
\end{align*}
The latent space feature difference $\epsilon(x,y,i,j)$ is the sum of squared residuals difference in the latent feature $h$ to a reference encoding $h_{\mathrm{reference}}$ at position $(i,j)$
\begin{align*}
\epsilon(x,y,i,j)=\sum (h(x,y)-h_{\mathrm{reference}}(i,j))^2,
\end{align*}

\subsubsection{Weighted DFE}

\begin{figure}[tb!]
 \centering
 \includegraphics[width=\linewidth]{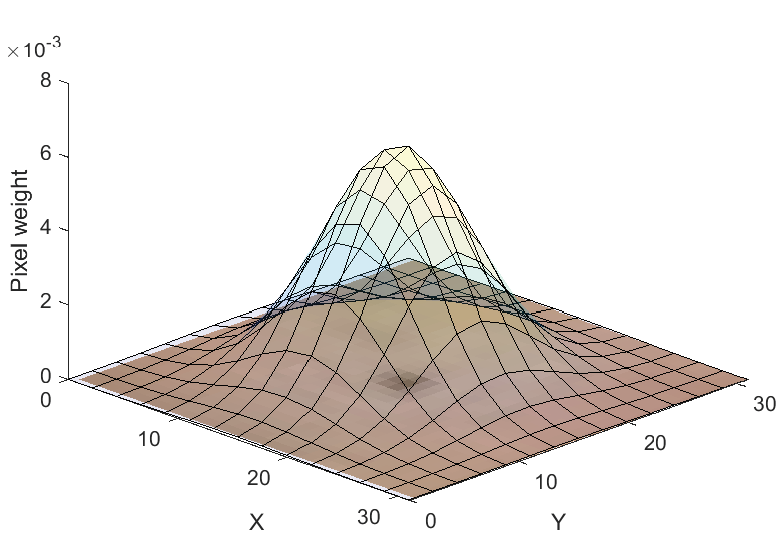}
\caption[Weights of the 2D Gaussian with a standard deviation of five.]{Weights of the 2D Gaussian with a standard deviation of five. We plotted the reference crop of the face mole under static conditions for easier visualisation.}
\label{fig:wdfe_gaussian_weight}
\end{figure}

To mitigate the influence of edge pixels on local $\epsilon$, we opted to retrain a weighted autoencoder version of DFE (wDFE). 
This decision stems from our observation that edge regions often exhibit drastic changes, whereas the model should prioritize the center of the image, where the reference feature is typically located. 
We achieved this weighting by incorporating a 2D Gaussian with a mean of 0 and a standard deviation of 5 into the mean squared error function.
Therefore, a weighted crop $C_{w}(m,n)$ with its origin at the center is:
\begin{align*}
	C_{w} (m,n)= \frac{(C_{p}(m,n)-C_{t}(m,n))^2}{2\pi\sigma^2}e^{(-\frac{m^2-n^2}{2\sigma^2})},
\end{align*}
where $C_{p}(m,n)$, $C_{t}(m,n)$, $\sigma$ represent the predicted crop, ground truth crop, and standard deviation of the Gaussian. 
A visualisation of the pixel weights for the reference crop of the face mole under static conditions is shown in Figure~\ref{fig:wdfe_gaussian_weight}.
This choice of standard deviation, representing a range within which 98.9\% of values fall within 15 pixels, assigns greater importance to center pixels over edge pixels, resulting in a network that prioritizes central regions.

Additionally, we retrained the model using full-resolution images from our validation dataset. 
This approach aims to tailor the model more closely to our application needs while preserving fine details and minimizing information loss. 
Moreover, training on high-resolution images can enhance the adaptability of the model to downsampling effects, fostering resilience to changes in resolution commonly encountered in real-world scenarios. 
By training on full-resolution data and evaluating on downsized images, the model learns robust and generalizable features that might be effective across varying resolutions, ultimately improving performance in diverse deployment scenarios.

\subsection{Subpixel level prediction}
\label{sec:subpixel_level_prediction}

SIFT, SURF, and our DFE method achieve pixel-level accuracy by selecting the pixel with the minimum SSR as their prediction. 
For a fair comparison with LK and PIPs++, subpixel-level accuracy is needed. 
We achieve this by fitting a surface to the SSR between the reference point and all points in the second image, and performing local interpolation around the pixel with the minimum SSR.

The SSR $\epsilon$ depends on image coordinates $x$ and $y$. 
Initially, a global search finds the minimum SSR for pixel-level prediction. 
Then, a second-order surface is fitted to a 3x3 point neighborhood around the global minimum. 
In cases of multiple minimum SSR points, the one with maximum curvature is chosen. 
Once the point with the lowest SSR, positive curvature, and positive second derivative $\epsilon_{xx}$ is found, subpixel coordinates are determined by solving for zero gradients in both directions.
For a more detailed description of our subpixel level prediction, we refer to our previous work \citep{chang2021skin}.

Due to its end-to-end architecture, Cotracker predicts points only at the pixel level. 
Thus, subpixel accuracy is used for all but Cotracker.

\subsection{Evaluation}
\label{sec:evaluation}

We assess feature tracking through two scenarios: feature matching and tracking, both relying solely on information from the initial frame without updating the reference feature.

Feature matching analyzes the occurrence of large errors in the dataset, depicted by presenting errors in decreasing order. 
However, tracking algorithms commonly exhibit divergence, wherein they begin tracking the wrong feature following a substantial error. 
Hence, tracking evaluation involves monitoring the cumulative sum of standardized errors over time.
In our previous work \citep{chang2021skin} we also investigated the tracking scheme where we allowed the reference feature to update to the prediction of the previous frame.

Errors are squared and standardized based on the variances in the $x$ and $y$ directions for each body part and motion condition. 
These standardized errors are summed cumulatively for each frame. 
Additionally, a 99\% confidence interval (CI) line is plotted using the inverse of the $\chi^2$ Cumulative Distribution Function (CDF) at a probability of 0.99, with $2n_{\mathrm{frame}}$ degrees of freedom, where $n_{\mathrm{frame}}$ represents the frame number.


\section{Results and Discussion}\label{sec:results}
\subsection{Feature matching accuracy}
\label{sec:feature_matching_accuracy}

DFE alone located the skin features in every frame.
It also had the smallest mean error tracking the face mole under bike conditions and hand mole under PT conditions, while being comparable to the best method in all other cases, see Table~\ref{tab:mean_errors}.
wDFE is the best method for tracking the nose tip under bike conditions with DFE being the only other method that did not diverge.
For the static cases, the DFE is still comparable with LK, and Cotracker which are the best methods for the nose tip and face mole; respectively.
SIFT with threshold and SURF did not perform well under any body part-motion combination.
Cotracker and PIPs++, which are the most modern methods did not perform well under bike conditions. 
It is worth noting that the maximum possible distance in all cases is 516.14 pixels, equaling the length of the diagonal of the images of size $420 \times 300$ pixels.

\begin{table*}[tb!]
\small
\centering
\caption[Mean errors (in pixels).]{Mean errors (in pixels) for feature matching using the same reference feature. The best results are highlighted in bold. SIFT* stands for SIFT with the nearest neighbors threshold. The frame count for the static, bike, and PT conditions was 260, 169, and 39 frames; respectively.}
\label{tab:mean_errors}
\begin{tabular}{C{2.3cm} C{1.0cm} C{1.2cm} C{1.1cm} C{1.0 cm} C{1.0cm} C{1.1cm}  C{1.5cm} C{1.3cm}}
\hline
 &SIFT & SIFT* & SURF & LK& DFE& wDFE& CoTracker& PIPs++  \\
\hline
Static face mole& 17.278 & 230.085& 93.843 &1.076 & 1.037& 1.093&\textbf{1.036}& 2.999\\
Static nose tip& 2.374 & 385.687& 48.134 & \textbf{2.157}& 2.470&  2.269&2.178& 2.558\\
Bike face mole& 16.329 & 216.249 & 74.934  & 4.684 & \textbf{0.899}& 0.920& 85.543& 48.087 \\
Bike nose tip& 64.120 & 421.995 & 116.632 & 11.310 &3.306&  \textbf{2.618} & 90.687& 32.657 \\
PT hand mole&17.286  & 227.195  &85.267  & 33.669  & \textbf{0.639}&  30.084&1.161& 7.573 \\
\hline
\end{tabular}
\end{table*}

\begin{figure*}[tb!]
 \centering
 \includegraphics[width=\textwidth]{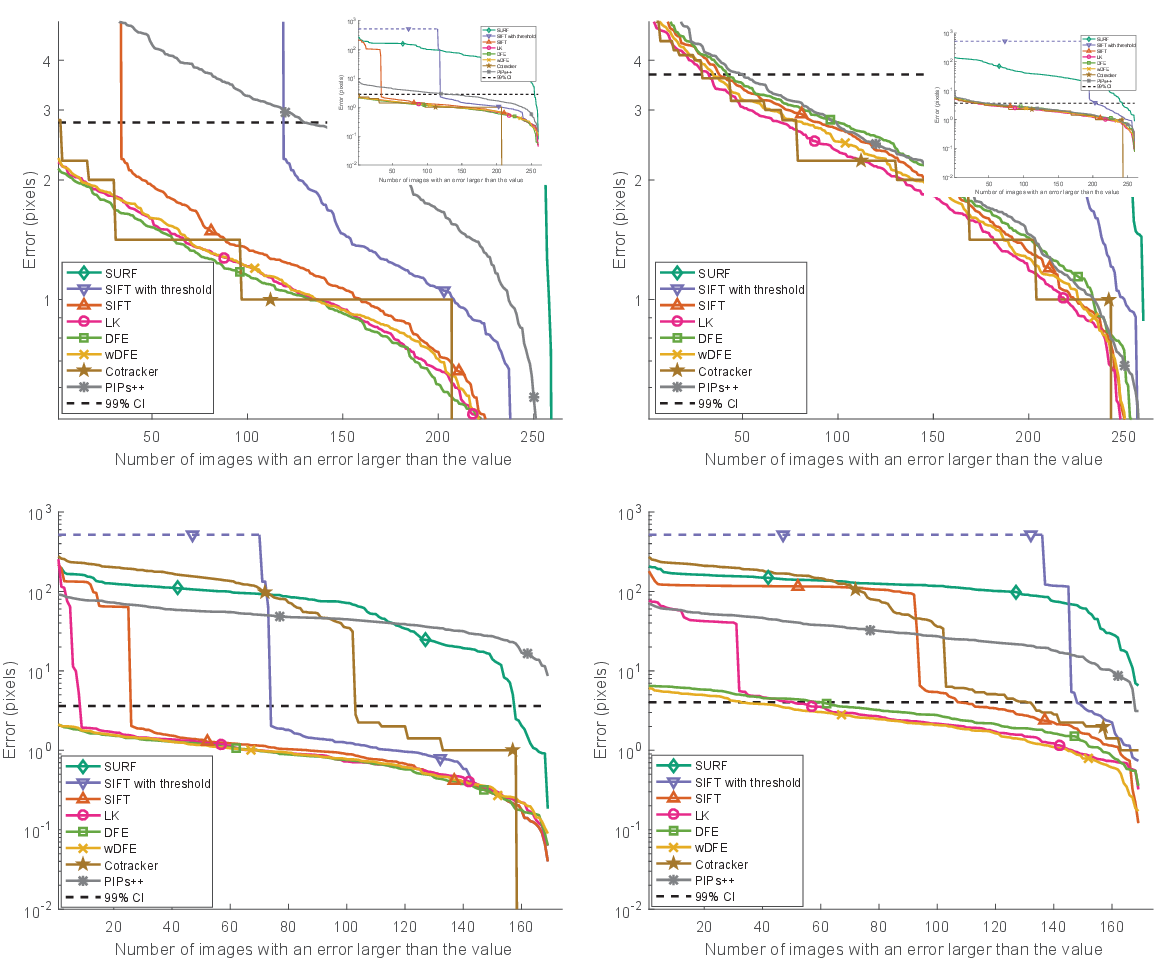}
\caption[Sorted errors for matching features across frames.]{Sorted errors for matching the face mole under static conditions (top-left), nose tip under static conditions (top-right), face mole under bike conditions (bottom-left), and nose tip under bike conditions (bottom-right). The purple dashed line for SIFT with threshold stands for the frames where the nearest neighbor distance threshold was larger than $0.8$. Since the predictions of Cotracker are at the pixel level, errors with 0 values are not visible in the log scale plot.}
\label{fig:sorted_matching_errors}
\end{figure*}

The distance errors (in pixels) of all methods sorted in descending order for each body part and motion condition are shown in Figure~\ref{fig:sorted_matching_errors}.
We note that LK, Cotracker, and PIPs++ are comparable to DFE for the static conditions.
The best performance for SIFT was while tracking the nose tip under static conditions.
SURF and SIFT with the recommended nearest neighbor to second nearest neighbor threshold did not perform well in any of the scenarios we tested.
Our DFE method was the only method to perform well in every condition.
Even while tracking the nose tip under bike conditions the error was at most $6.5$ pixels.
For the distinctive features, $i.e.$ moles, the error was at most $2.1$ pixels, while it exceeded $219$ for the other methods in some conditions. 

\subsection{Tracking}
\label{sec:tracking}

\begin{figure*}[tb!]
 \centering
  \includegraphics[width=\textwidth]{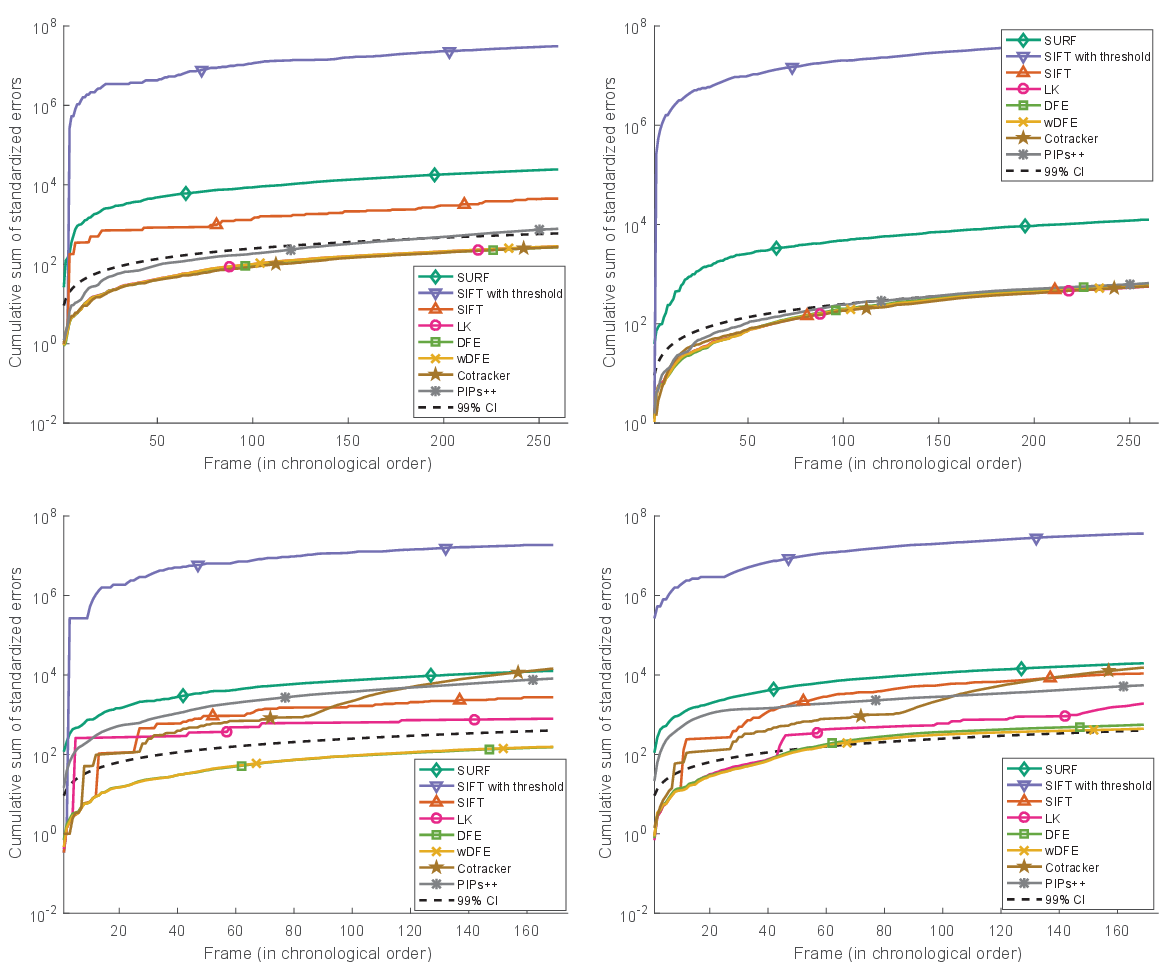}
\caption[Cumulative sum of standardized squared errors for tracking features using the reference feature from the original image.]{Cumulative sum of standardized squared errors using the reference feature from the original image for tracking the face mole under static conditions (top-left), nose tip under static conditions (top-right), face mole under bike conditions (bottom-left), and nose tip under bike conditions (bottom-right).}
\label{fig:tracking_errors_ofeat}
\end{figure*}

The results for tracking only the reference feature from the original image are shown in Figure~\ref{fig:tracking_errors_ofeat}.
The average errors using the reference feature from the original image are shown in Table~\ref{tab:mean_errors}.

On the tracking LK, CoTracker, PIPs++, wDFE, and DFE perform similarly under static conditions with CoTracker being superior on the face mole and LK being superior on the nose tip. 
Even though DFE is outperformed by other methods under static conditions, none of the errors are significantly different from each other since all are within the uncertainty of the manual labelling.
However, DFE outperformed all methods except wDFE for tracking under bike conditions.
Similarly to the static conditions, both methods are not statistically different from each other since all are within the uncertainty threshold.
The large errors for CoTracker for both body parts on this condition show that it failed to track them. 
Also, the cumulative sum of standardized squared errors of DFE even remains within the 99\% confidence interval for the distinctive skin mole. 

\subsection{SSR as a function of spatial coordinates}
\begin{figure*}[tb!]
 \centering
 \includegraphics[width=\textwidth]{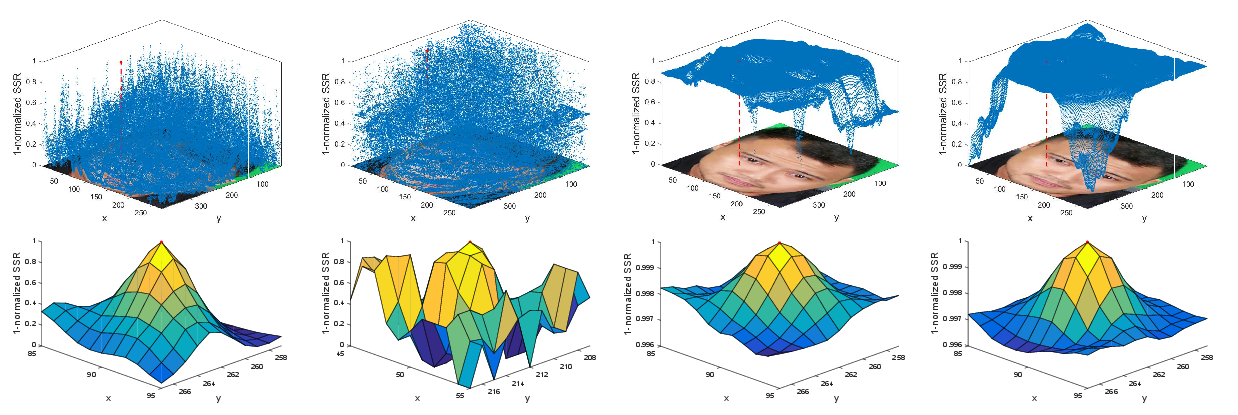}
\caption[SSR as a function of the spatial image coordinates.]{SSR as a function of the spatial image coordinates for the face mole under bike conditions for SIFT (left), SURF (middle-left), DFE (middle-right), and wDFE (right) and their corresponding regional enlargements of the optimum. The red dashed lines mark the point in the space with minimum SSR, which is plotted as a red circle.}
\label{fig:spatial_ssr_best}
\end{figure*}

We looked into the extreme cases of localisation performance of SIFT, SURF, DFE, and wDFE methods.
Figure~\ref{fig:spatial_ssr_best} shows the SSR landscape of the reference face mole under static conditions.
We can see that the SSR landscape of SIFT and SURF is chaotic and does not have any clear structure or pattern.
This is in contrast to the SSR landscape of DFE and wDFE which has smaller values near the face.
This suggests that our method is better at creating feature descriptors for skin features than the other traditional methods.
A structured SSR landscape will prevent the algorithm from selecting a false positive that is spatially far from the ground truth.

\subsection{Generalisation to PT data}
\begin{figure*}[tb!]
 \centering
  \includegraphics[width=\textwidth]{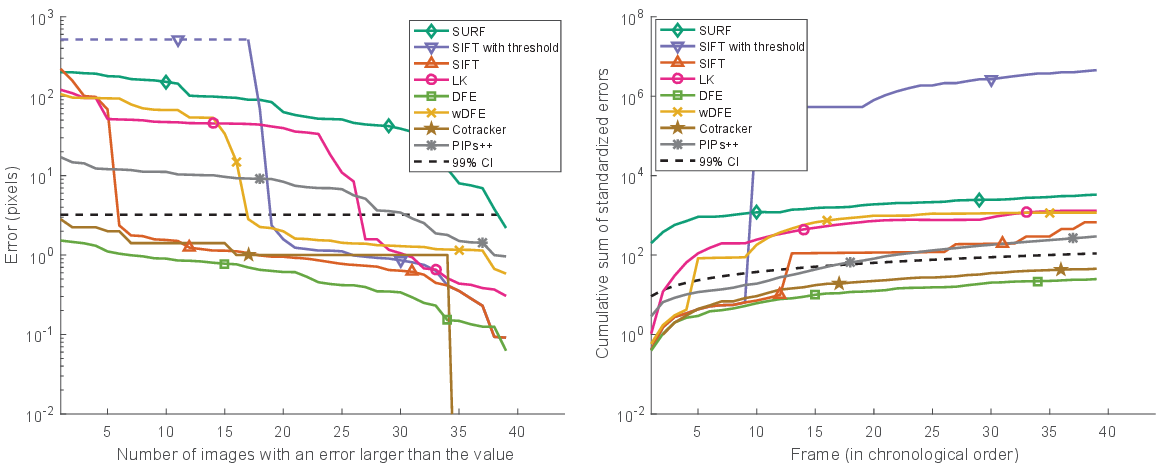}
\caption[Results from the methods for tracking the mole on the hand of the PD patient.]{Results of running the methods for tracking the mole on the hand of the PD patient. Sorted errors for tracking the mole (left) and cumulative error for tracking using the reference feature from the original image (right).}
\label{fig:results_pd}
\end{figure*}

The performance of the methods is qualitatively the same as for the face mole under bike conditions based on a comparison of Figures~\ref{fig:results_pd},~\ref{fig:sorted_matching_errors}, and ~\ref{fig:tracking_errors_ofeat}.
DFE is still the best method for tracking the mole on the hand of the PD patient.
In contrast to Figure~\ref{fig:sorted_matching_errors}, the second-best method is CoTracker, followed by PIPs++.
Here, wDFE did not perform as it diverged around frame 4, which caused it to have a large error. 
The images in this dataset are qualitatively similar and there is little distortion due to motion. 
This makes the DFE result qualitatively closer to the face mole under bike conditions than under static conditions.
SURF again failed to track the mole. 
These show that DFE is the best tracking method, no matter what tracking scheme is employed, as it was the only method that did not diverge.

\section{Conclusions}\label{sec:conclusions}

Our algorithm surpasses both the latest CoTracker and PIPs++, as well as traditional computer vision methods, in accuracy, and does not diverge for skin feature matching in ballistocardiography and PD applications. 
In this work, we have added the comparisons to PIPs++ and wDFE, which were not included in our previous work \citep{chang2021skin}.
Global minima, housing multiple nearest neighbors with similar SSR predictions, create a smoother SSR landscape compared to SIFT and SURF. 
This reduces false positives resulting from high-dimensional feature vector similarity calculations.
The distinctiveness of its feature representations stems from spatial information preservation through convolution, hierarchical feature learning for complex feature acquisition, and the creation of latent variables capturing sample variance. 
Experimentally, incorrect predictions by the matching method yield large errors, leading to divergence in tracking algorithms.

It is also possible that DFE takes advantage of the image being coded in the CIELAB space rather than the other learning-based methods that operate on RGB-coded images. 
In the CIELAB space, the distance between colors is pseudo-uniform, aligning more closely with how humans perceive color. 
When dealing with very low errors, such as a couple of pixels as in this case, this information might play a crucial role in differentiating between the methods. 

To ensure that our testing is exhaustive, yet simple and that large changes cannot cause trouble, we currently do not search locally for features but rather do a global search for the best match.
Thus, our method is not computationally optimized.
It currently takes approximately four times longer than SIFT, 27 times longer than PIPs++, and 42 times longer than CoTracker to run on the validation data with our hardware. 
We could assume that the skin features have approximate registration and perform a local search to speed up the computation in the future.
In addition, the matching scheme could be sped up by implementing an approximate technique for matching neighbors as has been demonstrated in SIFT.
The search could also be optimized if we took advantage of the convexity of the search space, which we have locally for DFE in all cases.
We have not considered the occlusion of the key point.

In scenarios where there is abundant image data but limited or absent labels, DFE presents an alternative solution. 
By leveraging the unsupervised nature of the autoencoder, DFE offers a superior solution compared to off-the-shelf methods for our applications. 
Its ability to extract meaningful features from unlabelled data makes it particularly effective in situations where PIPs++ and CoTracker may be impractical or infeasible. 
Most state-of-the-art methods for computer vision are supervised because they require labeled training data to learn the complex relationships between input images and their corresponding outputs, offering the advantage of guided learning and enabling the models to generalize effectively to new, unseen data, unlike unsupervised methods which often lack this guidance and may struggle with generalization.
Additionally, the adaptability of DFE to diverse datasets and its capacity to handle large volumes of unlabelled data make it a valuable tool for various image analysis tasks.

\section{Acknowledgments}

This work was supported by the Ministry of Science and Technology in Taiwan (MOST 108-2218-E-006-046, 109-2224-E-006-003, 111-2221-E-006-186, and 110-2222-E-006-010).
We thank Chien-Chih Wang for collecting the ballistocardiography data as a part of his Master's thesis work.
We thank Wei-Pei Shi for helping compare DFE to other state-of-the-art methods. 

\section{Compliance with Ethical Standards}
Our research involves Human Participants. 
The use of the UTKface and remote ballistocardiography datasets was reviewed and approved by the National Cheng Kung University Human Research Ethics Review Committee under case 108-244.
The use of the Parkinson's Disease postural tremor test dataset was reviewed and approved by the Kaohsiung Medical University Chung-Ho Memorial Hospital Institutional Review Board under the number KMUHIRB-E(I)-20190173.

\section{Statements and declarations}

\subsection{Funding}
This research was supported by the Ministry of Science and Technology in Taiwan (MOST 108-2218-E-006-046, 109-2224-E-006-003, and 111-2221-E-006-186).

\subsection{Competing interests}
The authors have no financial or non-financial interests that are directly or indirectly related to the work submitted for publication.

\subsection{Ethical and informed consent for data used}
The use of the UTKface and remote ballistocardiography datasets was reviewed and approved by the National Cheng Kung University Human Research Ethics Review Committee under case 108-244.
The use of the Parkinson's Disease postural tremor test dataset was reviewed and approved by the Kaohsiung Medical University Chung-Ho Memorial Hospital Institutional Review Board under the number KMUHIRB-E(I)-20190173. 
In both instances, the participants provided written, informed consent before the start of the experiment.

\subsection{Data availability and access}
Our ethics approval does not allow us to make the validation data publicly available. 
The training of the autoencoder used in the Deep Feature Encodings can be reproduced using the publicly available UTKface dataset \url{https://susanqq.github.io/UTKFace/}.

\subsection{Code availability}
Our model and code are available in the following repository: \url{https://bitbucket.org/nordlinglab/nordlinglab-dfe/src/main/}.

\subsection{Authors' contribution statement}
Author contribution using the CRediT taxonomy: 
Conceptualization: TN; Data curation: JC; Formal analysis: JC; Methodology: JC and TN; Investigation: JC; Software: JC; Verification: JC and TN; Visualization: JC and TN; Writing - original draft preparation: JC; Writing - review and editing: TN; Funding acquisition: TN; Project administration: TN; Resources: TN; Supervision: TN.

\clearpage

\begin{thebibliography}{40}
\providecommand{\natexlab}[1]{#1}
\providecommand{\url}[1]{\texttt{#1}}
\expandafter\ifx\csname urlstyle\endcsname\relax
  \providecommand{\doi}[1]{doi: #1}\else
  \providecommand{\doi}{doi: \begingroup \urlstyle{rm}\Url}\fi

\bibitem[Ashyani et~al.(2022)Ashyani, Lin, Roman, Yeh, Kuo, Tsai, Lin, Tu, Su,
  Wang, Tan, and Nordling]{ashyani2022digitization}
Akram Ashyani, Chi-Lun Lin, Esteban Roman, Ted Yeh, Tachyon Kuo, Wei-Fang Tsai,
  Yushan Lin, Ric Tu, Austin Su, Chien-Chih Wang, Chun-Hsiang Tan, and
  Torbj{\"o}rn E~M Nordling.
\newblock Digitization of updrs upper limb motor examinations towards automated
  quantification of symptoms of parkinson's disease.
\newblock \emph{Manuscript in preparation}, 2022.

\bibitem[Bay et~al.(2006)Bay, Tuytelaars, and Van~Gool]{Bay2006}
Herbert Bay, Tinne Tuytelaars, and Luc Van~Gool.
\newblock Surf: Speeded up robust features.
\newblock In \emph{European conference on computer vision}, pages 404--417.
  Springer, 2006.

\bibitem[Biggs et~al.(2019)Biggs, Roddick, Fitzgibbon, and
  Cipolla]{biggs2019creatures}
Benjamin Biggs, Thomas Roddick, Andrew Fitzgibbon, and Roberto Cipolla.
\newblock Creatures great and smal: Recovering the shape and motion of animals
  from video.
\newblock In \emph{Computer Vision--ACCV 2018: 14th Asian Conference on
  Computer Vision, Perth, Australia, December 2--6, 2018, Revised Selected
  Papers, Part V 14}, pages 3--19. Springer, 2019.

\bibitem[Butler et~al.(2012)Butler, Wulff, Stanley, and
  Black]{butler2012naturalistic}
Daniel~J Butler, Jonas Wulff, Garrett~B Stanley, and Michael~J Black.
\newblock A naturalistic open source movie for optical flow evaluation.
\newblock In \emph{European conference on computer vision}, pages 611--625.
  Springer, 2012.

\bibitem[Caron et~al.(2021)Caron, Touvron, Misra, J{\'e}gou, Mairal,
  Bojanowski, and Joulin]{caron2021emerging}
Mathilde Caron, Hugo Touvron, Ishan Misra, Herv{\'e} J{\'e}gou, Julien Mairal,
  Piotr Bojanowski, and Armand Joulin.
\newblock Emerging properties in self-supervised vision transformers.
\newblock In \emph{Proceedings of the IEEE/CVF international conference on
  computer vision}, pages 9650--9660, 2021.

\bibitem[Chang and Nordling(2021)]{chang2021skin}
Jose~Ramon Chang and Torbj{\"{o}}rn E.~M. Nordling.
\newblock Skin feature point tracking using deep feature encodings.
\newblock \emph{arXiv preprint}, dec 2021.
\newblock URL \url{https://arxiv.org/abs/2112.14159}.

\bibitem[Cheng et~al.(2021{\natexlab{a}})Cheng, Wong, Chin, Chan, and
  So]{cheng2021deep}
Chun-Hong Cheng, Kwan-Long Wong, Jing-Wei Chin, Tsz-Tai Chan, and Richard~HY
  So.
\newblock Deep learning methods for remote heart rate measurement: A review and
  future research agenda.
\newblock \emph{Sensors}, 21\penalty0 (18):\penalty0 6296, 2021{\natexlab{a}}.

\bibitem[Cheng et~al.(2021{\natexlab{b}})Cheng, Wang, Bao, and
  Lu]{cheng2021appearance}
Yihua Cheng, Haofei Wang, Yiwei Bao, and Feng Lu.
\newblock Appearance-based gaze estimation with deep learning: A review and
  benchmark.
\newblock \emph{arXiv preprint arXiv:2104.12668}, 2021{\natexlab{b}}.

\bibitem[Ciaparrone et~al.(2020)Ciaparrone, S{\'a}nchez, Tabik, Troiano,
  Tagliaferri, and Herrera]{Ciaparrone2020}
Gioele Ciaparrone, Francisco~Luque S{\'a}nchez, Siham Tabik, Luigi Troiano,
  Roberto Tagliaferri, and Francisco Herrera.
\newblock Deep learning in video multi-object tracking: A survey.
\newblock \emph{Neurocomputing}, 381:\penalty0 61--88, 2020.

\bibitem[Dai et~al.(2016)Dai, Li, He, and Sun]{Dai2016}
Jifeng Dai, Yi~Li, Kaiming He, and Jian Sun.
\newblock R-fcn: Object detection via region-based fully convolutional
  networks.
\newblock In \emph{Advances in neural information processing systems}, pages
  379--387, 2016.

\bibitem[Doersch et~al.(2022)Doersch, Gupta, Markeeva, Recasens, Smaira, Aytar,
  Carreira, Zisserman, and Yang]{doersch2022tap}
Carl Doersch, Ankush Gupta, Larisa Markeeva, Adri{\`a} Recasens, Lucas Smaira,
  Yusuf Aytar, Jo{\~a}o Carreira, Andrew Zisserman, and Yi~Yang.
\newblock Tap-vid: A benchmark for tracking any point in a video.
\newblock \emph{Advances in Neural Information Processing Systems},
  35:\penalty0 13610--13626, 2022.

\bibitem[Dosovitskiy et~al.(2015)Dosovitskiy, Fischer, Ilg, Hausser, Hazirbas,
  Golkov, Van Der~Smagt, Cremers, and Brox]{dosovitskiy2015flownet}
Alexey Dosovitskiy, Philipp Fischer, Eddy Ilg, Philip Hausser, Caner Hazirbas,
  Vladimir Golkov, Patrick Van Der~Smagt, Daniel Cremers, and Thomas Brox.
\newblock Flownet: Learning optical flow with convolutional networks.
\newblock In \emph{Proceedings of the IEEE international conference on computer
  vision}, pages 2758--2766, 2015.

\bibitem[Gaidon et~al.(2016)Gaidon, Wang, Cabon, and Vig]{gaidon2016virtual}
Adrien Gaidon, Qiao Wang, Yohann Cabon, and Eleonora Vig.
\newblock Virtual worlds as proxy for multi-object tracking analysis.
\newblock In \emph{Proceedings of the IEEE conference on computer vision and
  pattern recognition}, pages 4340--4349, 2016.

\bibitem[Goodfellow et~al.(2016)Goodfellow, Bengio, and
  Courville]{Goodfellow2016}
Ian Goodfellow, Yoshua Bengio, and Aaron Courville.
\newblock \emph{{Deep learning}}.
\newblock MIT Press, Cambridge, MA, U.S.A., 2016.
\newblock ISBN 978-0262035613.
\newblock URL \url{http://www.deeplearningbook.org}.

\bibitem[Harley et~al.(2022)Harley, Fang, and Fragkiadaki]{harley2022particle}
Adam~W Harley, Zhaoyuan Fang, and Katerina Fragkiadaki.
\newblock Particle video revisited: Tracking through occlusions using point
  trajectories.
\newblock In \emph{European Conference on Computer Vision}, pages 59--75.
  Springer, 2022.

\bibitem[{He} et~al.(2016){He}, {Zhang}, {Ren}, and {Sun}]{He2016deep}
K.~{He}, X.~{Zhang}, S.~{Ren}, and J.~{Sun}.
\newblock Deep residual learning for image recognition.
\newblock In \emph{2016 IEEE Conference on Computer Vision and Pattern
  Recognition (CVPR)}, pages 770--778, June 2016.
\newblock \doi{10.1109/CVPR.2016.90}.

\bibitem[He et~al.(2017)He, Gkioxari, Dollár, and Girshick]{He2017}
Kaiming He, Georgia Gkioxari, Piotr Dollár, and Ross Girshick.
\newblock Mask r-cnn.
\newblock In \emph{2017 IEEE International Conference on Computer Vision
  (ICCV)}, pages 2980--2988, 2017.
\newblock \doi{10.1109/ICCV.2017.322}.

\bibitem[Karaev et~al.(2023)Karaev, Rocco, Graham, Neverova, Vedaldi, and
  Rupprecht]{karaev2023cotracker}
Nikita Karaev, Ignacio Rocco, Benjamin Graham, Natalia Neverova, Andrea
  Vedaldi, and Christian Rupprecht.
\newblock Cotracker: It is better to track together.
\newblock \emph{arXiv preprint arXiv:2307.07635}, 2023.

\bibitem[Lee et~al.(2021)Lee, Seong, Ozlu, Shim, Marakhimov, and
  Lee]{lee2021biosignal}
Wookey Lee, Jessica~Jiwon Seong, Busra Ozlu, Bong~Sup Shim, Azizbek Marakhimov,
  and Suan Lee.
\newblock Biosignal sensors and deep learning-based speech recognition: A
  review.
\newblock \emph{Sensors}, 21\penalty0 (4):\penalty0 1399, 2021.

\bibitem[Liu et~al.(2016)Liu, Anguelov, Erhan, Szegedy, Reed, Fu, and
  Berg]{Liu2016ssd}
Wei Liu, Dragomir Anguelov, Dumitru Erhan, Christian Szegedy, Scott Reed,
  Cheng-Yang Fu, and Alexander~C Berg.
\newblock Ssd: Single shot multibox detector.
\newblock In \emph{European conference on computer vision}, pages 21--37.
  Springer, 2016.

\bibitem[Lowe(2004)]{Lowe2004}
David Lowe.
\newblock Distinctive image features from scale-invariant keypoints.
\newblock \emph{International Journal of Computer Vision}, 60:\penalty0 91--,
  11 2004.
\newblock \doi{10.1023/B:VISI.0000029664.99615.94}.

\bibitem[Lucas and Kanade(1981)]{lucas1981iterative}
Bruce Lucas and Takeo Kanade.
\newblock An iterative image registration technique with an application to
  stereo vision.
\newblock In \emph{IJCAI}, volume~81, 04 1981.

\bibitem[Manni et~al.(2020)Manni, van~der Sommen, Zinger, Shan, Holthuizen,
  Lai, Bustr{\"o}m, Hoveling, Edstr{\"o}m, Elmi-Terander,
  et~al.]{manni2020hyperspectral}
Francesca Manni, Fons van~der Sommen, Svitlana Zinger, Caifeng Shan, Ronald
  Holthuizen, Marco Lai, Gustav Bustr{\"o}m, Richelle~JM Hoveling, Erik
  Edstr{\"o}m, Adrian Elmi-Terander, et~al.
\newblock Hyperspectral imaging for skin feature detection: Advances in
  markerless tracking for spine surgery.
\newblock \emph{Applied Sciences}, 10\penalty0 (12):\penalty0 4078, 2020.

\bibitem[Mayer et~al.(2016)Mayer, Ilg, Hausser, Fischer, Cremers, Dosovitskiy,
  and Brox]{mayer2016large}
Nikolaus Mayer, Eddy Ilg, Philip Hausser, Philipp Fischer, Daniel Cremers,
  Alexey Dosovitskiy, and Thomas Brox.
\newblock A large dataset to train convolutional networks for disparity,
  optical flow, and scene flow estimation.
\newblock In \emph{Proceedings of the IEEE conference on computer vision and
  pattern recognition}, pages 4040--4048, 2016.

\bibitem[McLaren(1976)]{mclaren1976xiii}
K~McLaren.
\newblock Xiii--the development of the cie 1976 (l* a* b*) uniform colour space
  and colour-difference formula.
\newblock \emph{J. of the Soc. of Dyers and Colour.}, 92\penalty0 (9):\penalty0
  338--341, 1976.

\bibitem[Ni et~al.(2021)Ni, Azarang, and Kehtarnavaz]{ni2021review}
Aoxin Ni, Arian Azarang, and Nasser Kehtarnavaz.
\newblock A review of deep learning-based contactless heart rate measurement
  methods.
\newblock \emph{Sensors}, 21\penalty0 (11):\penalty0 3719, 2021.

\bibitem[Noh et~al.(2017)Noh, Araujo, Sim, Weyand, and Han]{noh2017large}
Hyeonwoo Noh, Andre Araujo, Jack Sim, Tobias Weyand, and Bohyung Han.
\newblock Large-scale image retrieval with attentive deep local features.
\newblock In \emph{Proceedings of the IEEE international conference on computer
  vision}, pages 3456--3465, 2017.

\bibitem[Reis et~al.(2023)Reis, Kupec, Hong, and Daoudi]{reis2023real}
Dillon Reis, Jordan Kupec, Jacqueline Hong, and Ahmad Daoudi.
\newblock Real-time flying object detection with yolov8.
\newblock \emph{arXiv preprint arXiv:2305.09972}, 2023.

\bibitem[Ren et~al.(2015)Ren, He, Girshick, and Sun]{Ren2015}
Shaoqing Ren, Kaiming He, Ross Girshick, and Jian Sun.
\newblock Faster r-cnn: Towards real-time object detection with region proposal
  networks.
\newblock In \emph{Advances in neural information processing systems}, pages
  91--99, 2015.

\bibitem[Rocco et~al.(2023)Rocco, Makarov, Kokkinos, Novotny, Graham, Neverova,
  and Vedaldi]{rocco2023real}
Ignacio Rocco, Iurii Makarov, Filippos Kokkinos, David Novotny, Benjamin
  Graham, Natalia Neverova, and Andrea Vedaldi.
\newblock Real-time volumetric rendering of dynamic humans.
\newblock \emph{arXiv preprint arXiv:2303.11898}, 2023.

\bibitem[Sikander and Anwar(2018)]{sikander2018driver}
Gulbadan Sikander and Shahzad Anwar.
\newblock Driver fatigue detection systems: A review.
\newblock \emph{IEEE Transactions on Intelligent Transportation Systems},
  20\penalty0 (6):\penalty0 2339--2352, 2018.

\bibitem[Stofa et~al.(2021)Stofa, Zulkifley, and Zainuri]{stofa2021skin}
Marzuraikah~Mohd Stofa, Mohd~Asyraf Zulkifley, and Muhammad Ammirrul Atiqi~Mohd
  Zainuri.
\newblock Skin lesions classification and segmentation: A review.
\newblock \emph{International Journal of Advanced Computer Science and
  Applications}, 12\penalty0 (10), 2021.

\bibitem[Sun et~al.(2021)Sun, Shen, Wang, Bao, and Zhou]{sun2021loftr}
Jiaming Sun, Zehong Shen, Yuang Wang, Hujun Bao, and Xiaowei Zhou.
\newblock Loftr: Detector-free local feature matching with transformers.
\newblock In \emph{Proceedings of the IEEE/CVF conference on computer vision
  and pattern recognition}, pages 8922--8931, 2021.

\bibitem[Sundararaman et~al.(2021)Sundararaman, De~Almeida~Braga, Marchand, and
  Pettre]{sundararaman2021tracking}
Ramana Sundararaman, Cedric De~Almeida~Braga, Eric Marchand, and Julien Pettre.
\newblock Tracking pedestrian heads in dense crowd.
\newblock In \emph{Proceedings of the IEEE/CVF conference on computer vision
  and pattern recognition}, pages 3865--3875, 2021.

\bibitem[Teed and Deng(2020)]{teed2020raft}
Zachary Teed and Jia Deng.
\newblock Raft: Recurrent all-pairs field transforms for optical flow.
\newblock In \emph{Computer Vision--ECCV 2020: 16th European Conference,
  Glasgow, UK, August 23--28, 2020, Proceedings, Part II 16}, pages 402--419.
  Springer, 2020.

\bibitem[Vivaldy et~al.(2023)Vivaldy, Wang, Meher, and Nordling]{Vivaldy2023}
Gavin Vivaldy, Chien-Chin Wang, Jagmohan Meher, and Torbj{\"{o}}rn E.~M.
  Nordling.
\newblock {Protocol for collection of synchronised facial video,
  Electrocardiography, and Photoplethysmography data for remote
  Photoplethysmography model training and evaluation}.
\newblock \emph{Manuscript in preparation}, 2023.

\bibitem[Wang et~al.(2018)Wang, Gao, Tao, Yang, and Li]{Wang2018}
Nannan Wang, Xinbo Gao, Dacheng Tao, Heng Yang, and Xuelong Li.
\newblock Facial feature point detection.
\newblock \emph{Neurocomput.}, 275\penalty0 (C):\penalty0 50–65, January
  2018.
\newblock ISSN 0925-2312.
\newblock \doi{10.1016/j.neucom.2017.05.013}.
\newblock URL \url{https://doi.org/10.1016/j.neucom.2017.05.013}.

\bibitem[Weinzaepfel et~al.(2013)Weinzaepfel, Revaud, Harchaoui, and
  Schmid]{weinzaepfel2013deepflow}
Philippe Weinzaepfel, Jerome Revaud, Zaid Harchaoui, and Cordelia Schmid.
\newblock Deepflow: Large displacement optical flow with deep matching.
\newblock In \emph{Proceedings of the IEEE international conference on computer
  vision}, pages 1385--1392, 2013.

\bibitem[Zhang et~al.(2017)Zhang, Song, and Qi]{Zhang2017age}
Zhifei Zhang, Yang Song, and Hairong Qi.
\newblock Age progression/regression by conditional adversarial autoencoder.
\newblock In \emph{IEEE Conference on Computer Vision and Pattern Recognition
  (CVPR)}. IEEE, 2017.

\bibitem[Zheng et~al.(2023)Zheng, Harley, Shen, Wetzstein, and
  Guibas]{zheng2023pointodyssey}
Yang Zheng, Adam~W Harley, Bokui Shen, Gordon Wetzstein, and Leonidas~J Guibas.
\newblock Pointodyssey: A large-scale synthetic dataset for long-term point
  tracking.
\newblock In \emph{Proceedings of the IEEE/CVF International Conference on
  Computer Vision}, pages 19855--19865, 2023.

\end{thebibliography}


\end{document}